\documentclass[sigconf]{acmart}
\usepackage{multirow, bm}
\usepackage{enumitem}
\usepackage{balance}
\usepackage{color}
\definecolor{url}{RGB}{250, 90, 229}

\AtBeginDocument{%
  }
\copyrightyear{2023}
\acmYear{2023}
\setcopyright{acmlicensed}

\acmConference[MM '23] {Proceedings of the 31st ACM International Conference on Multimedia}{October 29--November 3, 2023}{Ottawa, ON, Canada.}
\acmBooktitle{Proceedings of the 31st ACM International Conference on Multimedia (MM '23), October 29--November 3, 2023, Ottawa, ON, Canada}

\acmPrice{15.00}
\acmISBN{979-8-4007-0108-5/23/10}
\acmDOI{10.1145/3581783.3612191}

\begin{document}

\title{Guided Image Synthesis via Initial Image Editing \\ in Diffusion Model}
\renewcommand{\shorttitle}{Guided Image Synthesis via Initial Image Editing in Diffusion Model}

\author{Jiafeng Mao}
\affiliation{%
  \institution{The University of Tokyo}
  \city{Tokyo}
  \country{Japan}}
\email{mao@hal.t.u-tokyo.ac.jp}

\author{Xueting Wang}
\affiliation{%
  \institution{CyberAgent, Inc.}
  \city{Tokyo}
  \country{Japan}}
\email{wang_xueting@cyberagent.co.jp}

\author{Kiyoharu Aizawa}
\affiliation{%
  \institution{The University of Tokyo}
  \city{Tokyo}
  \country{Japan}}
\email{aizawa@hal.t.u-tokyo.ac.jp}

\renewcommand{\shortauthors}{Jiafeng Mao, Xueting Wang, \& Kiyoharu Aizawa}

\begin{abstract}
Diffusion models have the ability to generate high quality images by denoising pure Gaussian noise images. While previous research has primarily focused on improving the control of image generation through adjusting the denoising process, we propose a novel direction of manipulating the initial noise to control the generated image. Through experiments on stable diffusion, we show that blocks of pixels in the initial latent images have a preference for generating specific content, and that modifying these blocks can significantly influence the generated image. In particular, we show that modifying a part of the initial image affects the corresponding region of the generated image while leaving other regions unaffected, which is useful for repainting tasks. Furthermore, we find that the generation preferences of pixel blocks are primarily determined by their values, rather than their position. By moving pixel blocks with a tendency to generate user-desired content to user-specified regions, our approach achieves state-of-the-art performance in layout-to-image generation. Our results highlight the flexibility and power of initial image manipulation in controlling the generated image. \textcolor{url}{Project Page: \url{https://ut-mao.github.io/swap.github.io}}
\end{abstract}
\vspace{-10pt}

\vspace{-10pt}
\vspace{-20pt}

\begin{teaserfigure}
  \vspace{-10pt}
  \includegraphics[width=0.9\textwidth]{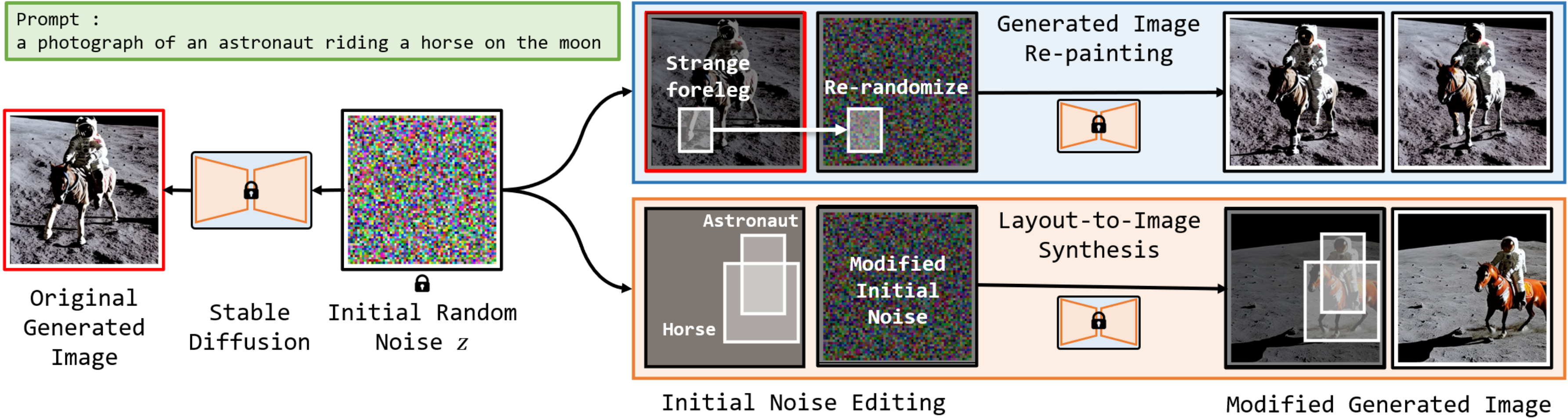}
  \vspace{-10pt}
  \caption{In this study, we investigate the impact of the initial image on image generation and propose a novel direction for controlling the generation process by manipulating the initial random noise. We present two direct applications of our findings: generated image re-painting and layout-to-image synthesis. Generated image re-painting allows users to modify a part of the generated image while preserving most of it. Layout-to-image synthesis requires generating objects in user-specified regions.}
  \label{fig:teaser}
\end{teaserfigure}

\maketitle

\begin{figure*}[t]
  \centering
  \includegraphics[width=0.9\linewidth]{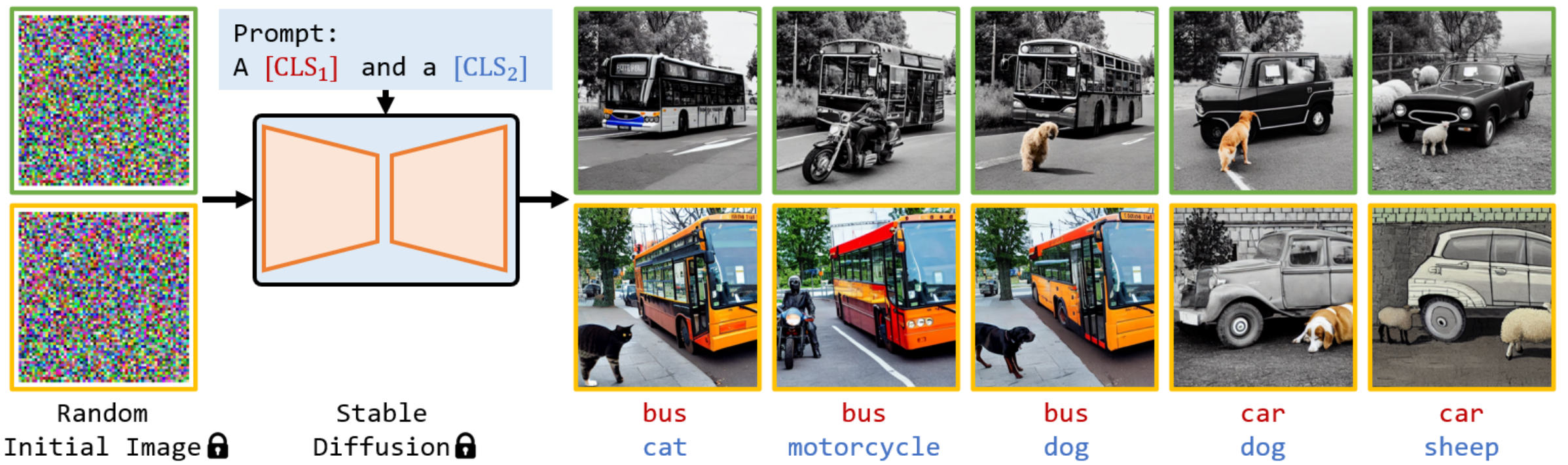}
  \vspace{-10pt}
  \caption{Our analysis experiments focused on the generation tendency of random initial noise Image. We present samples of generation results using two different initial images, both of which are randomly sampled from Gaussian distribution. Images with the same color borders are generated from the same initial image. When the initial images are the same, the same categories under different prompts are highly similar in position and visual appearance. When the randomly sampled initial images differ, the generated results are highly different in layout and details, even when the same prompts are used.}
  \label{fig:m1}
  \vspace{-15pt}
\end{figure*}

\vspace{-10pt}
\section{Introduction}

Image generation is an important area of research in computer vision. Diffusion models~\cite{ho2020denoising} have shown great potential for generating high-quality images by progressively denoising pure Gaussian noise images. Previous research~\cite{rombach2022high, ramesh2022hierarchical, nichol2022glide, kim2022diffusionclip, ramesh2021zero, ding2021cogview} in this area has primarily focused on using prompts to guide the generation of images. The prompt is usually a sentence describing the desired content of the generated image. The model then attempts to generate an image that matches the prompt's description. However, prompt-based text-to-image generation is highly uncertain and frequently fails. Although some methods~\cite{liu2022compositional, kawar2022imagic, feng2022training, hertz2022prompt, park2021benchmark, balaji2022ediffi} have attempted to achieve more controllable generation processes by improving the denoising process or exploring better ways to use prompt, in practical applications, users usually need to randomly generate dozens or even hundreds of different images using the same prompt before obtaining a satisfactory one. 

In this paper\footnote{This work was conducted while the first author was doing internship at CyberAgent, Inc.}, we investigate the critical reasons why the generation sometimes fails and further apply our conclusions to propose a new direction for controlling the generated contents. We notice that current widely used sampling methods, i.e., DDIM (denoising diffusion implicit models~\cite{song2020denoising}) and PLMS (pseudo linear multi-step method~\cite{liu2022pseudo}), are deterministic. Given a diffusion model $\theta$, prompt $p$ and a initial noise image $x_T$, the generated image $x_0$ is constant and can be represented as $x_0 = f_\theta(x_T, p)$. With the same prompt $p$ and model parameters $\theta$, the final generated image $x_0$ should be entirely determined by the randomly sampled initial image $x_T$. With such a critical position, the initial noise image has received little attention. This conclusion leads us to further analyze the initial image's impact on the generated results. 

Specifically, our experiments on stable diffusion demonstrated that the initial noise image has a preference for generating specific content, which we refer to as \textbf{generation tendency} in this paper. Specifically, once the initial image is determined, the generated images tend to exhibit significant similarities, such as layout, background, and style, even when different prompts are utilized. Additionally, we found that the generation tendency of each area of the initial image is relatively independent. Given a constant prompt, modifying the pixels in a single area of the initial image can result in a change in the corresponding area of generated image, while retaining the content of other areas. This finding can be applied to repainting tasks, where users are satisfied with most of the content in the generated image and wish to change a small portion of inappropriate content (top row in Fig.~\ref{fig:teaser}). We also discovered that the generation tendencies of regions on initial image are primarily related to their pixel values, rather than their spatial position. A region that tends to generate specific content is likely to generate the same content when moved to a different spatial location in the initial image. Therefore, simply moving all regions with the same tendency to generate specific content to a designated area can lead the diffusion model to generate the content in that area. This discovery can naturally be applied to the layout-to-image synthesis, which requires generating objects mentioned in the prompt within user-specified areas (bottom row in Fig.~\ref{fig:teaser}). Our method of modifying the initial image achieves a comparable performance with state-of-the-art methods when used alone, and can be combined with them to further improve control over the object's position.

By leveraging the initial image as a control mechanism, we can achieve fine-grained control over the generated images rather than relying solely on prompts or the denoising process. Our findings show that initial image manipulation is a flexible and powerful way to control the image generation of diffusion models, and open new avenues for research in image generation. We summarize the contributions of this paper as follows:

\begin{itemize}[leftmargin=*]
    \item Our experimental results indicate that the initial noise images exhibit distinct preferences in generating certain content, which are primarily determined by their pixel values. In other words, the initial image has a significant impact on the generated results.
    \item We identify a key reason why text-to-image generation sometimes fails.
    \item We propose a new direction for controlling the generation, which is utilizing the initial image as a control mechanism. Our ap- proach enables the repainting of generated images and achieves state-of-the-art performance on training-free layout-to-image generation tasks.
\end{itemize}
\vspace{-10pt}

\begin{figure*}[t]
  \centering
  \includegraphics[width=0.8\linewidth]{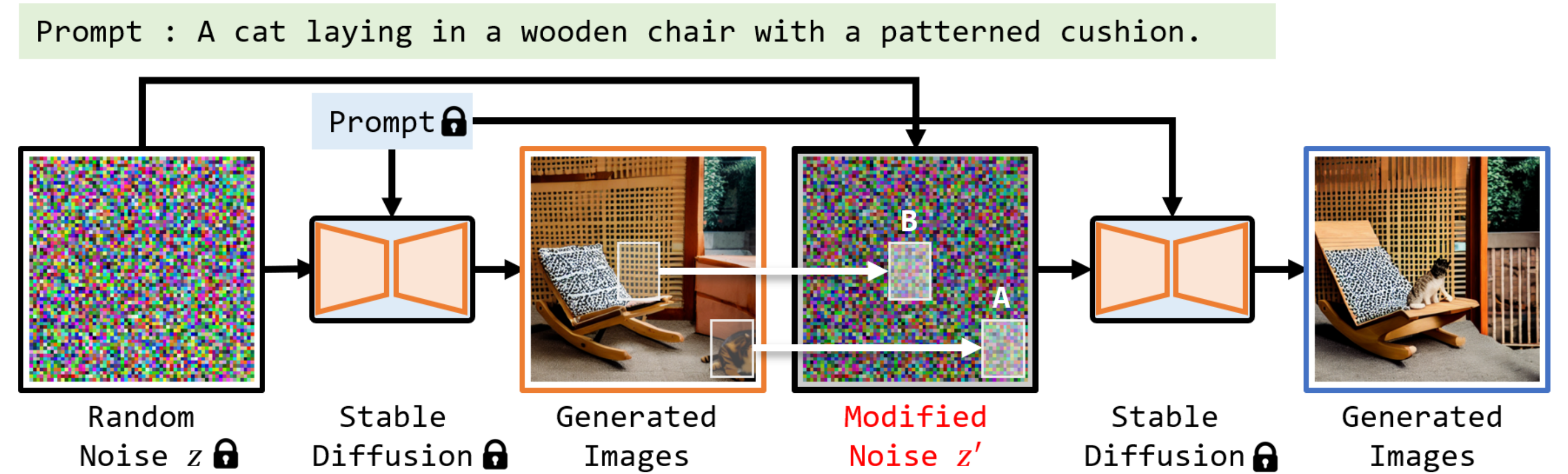}
  \vspace{-10pt}
  \caption{Our experiments on initial noise image editing. We first perform generation using random noise initial image sampled from Gaussian noise. Although both the 'cat' and the 'chair' are generated in the first generated image, the location of the cat is not satisfying to the prompt. Since the region where the cat is generated (bounding box A) that tends to generate 'cat' is not expected, we re-randomize the pixel values of the corresponding area in the initial image. We also re-randomize the region near the chair (bounding box B) because it did not tend to generate 'cat'. The improved image on the right side is generated by the modified initial image and the same prompt after a few attempts.}
  \label{fig:m2}
  \vspace{-15pt}
\end{figure*}

\section{Related Works}
\subsection{Diffusion Models}
Recent works~\cite{ho2020denoising, song2020denoising, nichol2021improved} on diffusion model have been shown to have the ability to produce samples that are comparable to real images. DDPM (Denoising Diffusion Probabilistic Models~\cite{ho2020denoising}) starts from pure Gaussian noise and keeps repeating the process of predicting the next distribution and randomly sampling new samples, eventually generating high-quality samples similar to actual pictures. However, DDPM requires hundreds to thousands of iterations to produce the final samples. Some previous works have successfully accelerated DDPM by tuning the variance schemes (improved DDPM~\cite{nichol2021improved}) or denoising equations (DDIM~\cite{song2020denoising}). DDIM relying on a non-Markovian process, accelerate the denoising process by taking multiple steps every iteration and replacing the random generation process with a deterministic one. Liu et al.~\cite{liu2022pseudo} extended DDIM into PNDMs and proposed a pseudo linear multi-step method (PLMS), achieving the fastest generation under similar generated quality.
Previous research has proposed various methods for controlling the generated content. Classifier guidance~\cite{dhariwal2021diffusion} denoises the image while updating it in the direction of high scores of the corresponding class based on the gradient of the classifier. Classifier-free~\cite{ho2021classifier} guidance achieves the same effect without a classifier. Most current methods~\cite{kim2022diffusionclip, ramesh2021zero, ding2021cogview, gafni2022make} take prompt guidance and calculate an updating direction by CLIP~\cite{radford2021learning} to guide the denoising. Stable diffusion~\cite{rombach2022high} denoises in the latent space and then decodes the latent to an image, which is currently the most widely used method. It augments the U-Net~\cite{ronneberger2015u} backbone with the cross-attention mechanism~\cite{vaswani2017attention} to enable the flexible form of a condition such as text. 

\vspace{-10pt}
\subsection{Fine-grained Controlling}
Generation solely based on text guidance can sometimes be difficult to achieve satisfactory results, as fine-grained control (e.g., size and location of objects, image style) may be difficult to express through text, and the generated results may not always match the description in the text. On the other hand, images generated by diffusion models have high uncertainty. Even with the same prompt, results can vary significantly each time. Therefore, some studies~\cite{liu2022compositional, kawar2022imagic, park2021benchmark, avrahami2022spatext} have attempted to achieve a more controllable generation process to improve the usability of diffusion models. Composable Diffusion~\cite{liu2022compositional} aims to achieve flexible control over image generation by taking in multiple text prompts. Specifically, it uses the sum of the guidance vectors corresponding to these prompts as the denoising direction. T2I-Adapter~\cite{mou2023t2i} uses adapters to provide extra guidance and achieve rich control and editing effects in the color and structure of the generation results. Some methods~\cite{zhang2023adding, huang2023composer, voynov2022sketch} attempt to achieve better controlling by improving models parameters. PITI~\cite{wang2022pretraining} proposes to provide structural guidance by closing the distance between the feature of other types of conditions and the text condition. Sketch-Guided~\cite{voynov2022sketch} proposes to utilize the similarity gradient between the target sketch and intermediate result to constrain the structure of the final results. It is observed that the cross-attention maps highly correspond to the layout and structure of the generated images~\cite{hertz2022prompt,feng2022training}, thus, several studies attempt to achieve fine-grained control through the manipulation of the attention maps. Structured Diffusion Guidance~\cite{feng2022training} proposes to obtain better CLIP embedding by analyzing the text's semantic structure in advance and replacing the attention map of certain words. Overall, these works focus on improving the controllability of the diffusion model by improving the generation process and the usage of prompt, ignoring the role of the initial image, which is our focus in this paper.

\vspace{-10pt}
\subsection{Layout-to-image synthesis}
Layout-to-image synthesis attempt to achieve more controllable generation by introducing object-wise guidance. Other than prompt, layout-to-image synthesis takes the specified location and size of objects described in the prompt. The generative model is expected to generate images where each object is positioned in the location specified by users. Some research~\cite{balaji2022ediffi, mao2023trainingfree} propose adding masks to the attention maps to increase the chances of generating specified objects in designated areas in order to achieve layout-to-image synthesis. These methods are fundamentally limited by the fact that they forcibly add features to the user-specified regions, completely ignoring the generation tendency of the initial image. In contrast, we use the initial image as a control mechanism and guide the diffusion model to proactively generate the user-specified content in the intended region.

\begin{figure*}[t]
  \centering
  \includegraphics[width=\linewidth]{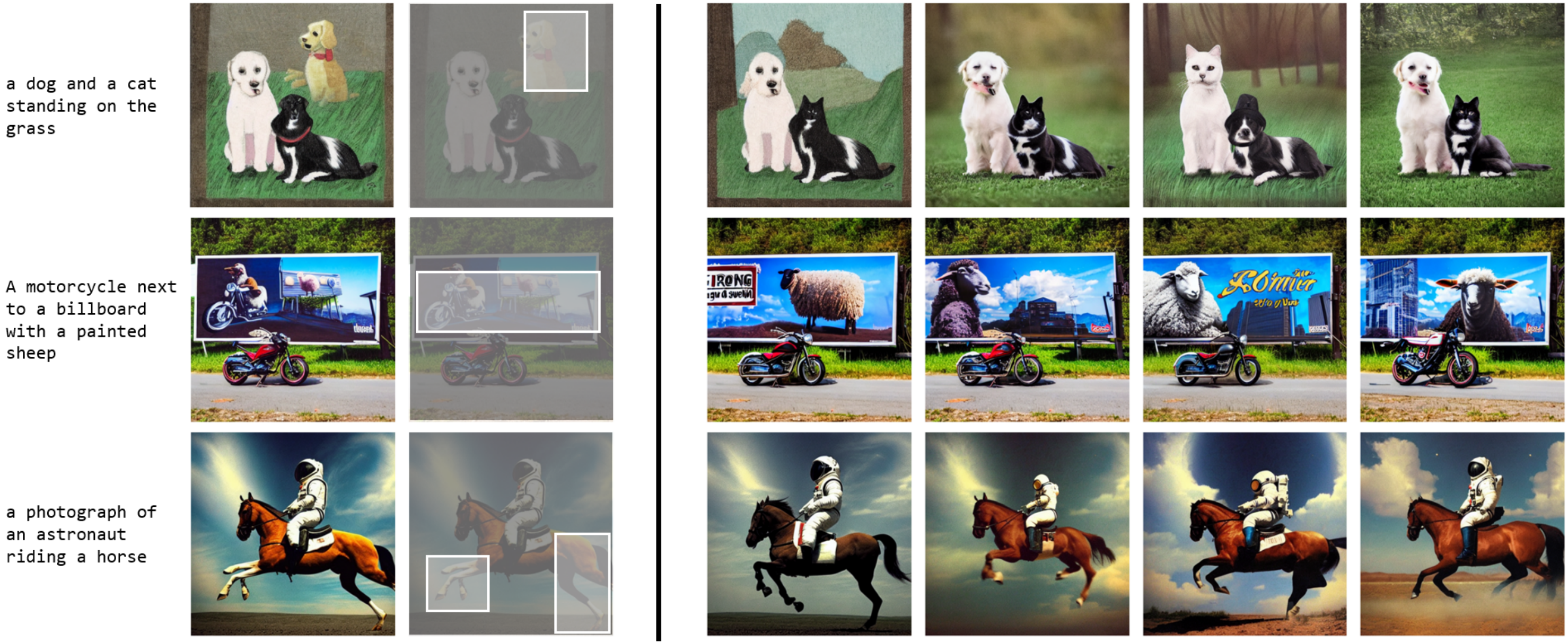}
  \vspace{-20pt}
  \caption{Samples of re-painting experiments. Images on the left side are generated from the random initial noise, while images on the right side are generated from partially re-sampled initial noise. The first prompt involves two objects, but three were generated. After re-sampling the pixel values in the region corresponding to the third object in the initial image, the third object was removed. The second generation failed to generate the billboard's content (which should be a sheep). After re-sampling the pixel values in the region corresponding to the billboard, we obtained images with the same layout, style, and correctly generated billboard. The horse in the third generated image has three forelegs and unnatural hind legs. After re-sampling these two parts, we obtained generated images with different and better horses while keeping the overall image unchanged. It takes 4-5 attempts on average to obtain each modified sample.}
  \label{fig:task_1}
  \vspace{-10pt}
\end{figure*}

\vspace{-10pt}
\section{Initial Image Editing}
In this section, we introduce our findings and proposals. We present experiments that discover the role of the initial image in the diffusion models for image generation in Sec.~\ref{sec:exp1}, which shows that different regions of the initial image have inherent tendencies towards generating certain contents. We analyze this phenomenon and provide a critical reason why text-to-image generation sometimes fails in Sec.~\ref{sec:ana}. We further attempt to manipulate the initial image to edit the generated images, making our finding applicable for image repainting tasks (Sec.~\ref{sec:exp2}) and layout-to-image synthesis tasks (Sec.~\ref{sec:exp3}).

\vspace{-10pt}
\subsection{Generation Tendency of Initial Image}
\label{sec:exp1}
We first conduct experiments to check the generation tendency of initial images. Specifically, we create a list of categories and then select two different categories, $\text{CLS}_1$ and $\text{CLS}_2$, from it to construct prompts for generation. The prompt takes the form of \verb|[a |$\text{CLS}_1$ \verb|and a| $\text{CLS}_2$\verb|]|~(e.g. \verb|[a dog and a car]|). As shown in Fig.~\ref{fig:m1}, we use fixed random initial images and a model with fixed parameters to generate each constructed prompt and observe the generation performance of the same initialization image guided by different prompts. When the same object is mentioned in different prompts, there is a high probability that this object will be generated in the same location and have a very similar visual appearance. For example, \verb|[a bus and a cat]|, \verb|[a bus and a motorcycle]|, \verb|[a bus and a dog]| all contain the word \verb|'bus'|, buses generated using the same initial image but different prompts appear similar. However, with different initial noise images (the bottom line), the construction of the generated images can vary greatly, even with the same prompt. Based on this experiment, we found that the initialized images themselves carry certain generation tendencies. Once the initialized image is determined, its generative tendency becomes fixed and affects the final generated image in conjunction with the prompt.

\vspace{-10pt}
\subsection{Analysis and Explanation}
\label{sec:ana}
In this section, we provide an analysis and explanation of the generation tendency of the initial image.

\noindent\textbf{Deterministic Denoising in Stable Diffusion} Stable diffusion~\cite{rombach2022high} can be interpreted as a sequence of denoising autoencoders $\epsilon_\theta(z_t, t)$, $t = 1, 2,.., T$, which are trained to predict added noise on input $z_t$, where $z_t$ is noised $z$. To generate the latent of an actual image, a stable diffusion model start from a pure Gaussian noise $z_T$ and denoises the latent image iteratively. Stable diffusion augments the U-Net~\cite{ronneberger2015u} backbone with the cross-attention mechanism~\cite{vaswani2017attention} to enable the flexible form of a condition such as text. For a text-to-image generation, a CLIP is employed to project condition $p$ to an intermediate representation $\tau_\theta(p)$, which is subsequently mapped to the intermediate layers of the U-Net through a cross-attention layer as follows, 

\begin{equation}
    \text{Attention}(Q,K,V) = \text{softmax}(\frac{QK^T}{\sqrt{d}})\cdot{V}, 
    \label{eq:attention}
\end{equation}

where ${Q=W^{(i)}_Q\cdot{z_t}}, ~K=W^{(i)}_K\cdot\tau_\theta(p),~V=W^{(i)}_V\cdot\tau_\theta(p)$. Thus, given a trained model with parameters $(W, \theta)$, the calculation of cross-attention map $\text{softmax}(\frac{QK^T}{\sqrt{d}})$ entirely depends on the initial latent $z_t$ and user-specified prompt $p$. Since the denoising sampling DDIM is also deterministic, once the initial random noise $z_t$ and the prompt $p$ is given, the predicted noise $\epsilon_\theta(z_t, t)$ of each step $t$ and the final denoised latent image $z_0$ are both determined.

\noindent\textbf{Cross-Attention Map} It has been stated that each entry of the cross-attention map $M^{(i)}$ corresponds to a certain word $p^{(i)}$ in the prompt $p$~\cite{hertz2022prompt,feng2022training}. Additionally, the layout and shape of objects in generated images are highly correlated with their attention maps. In simple terms, the cross-attention layer computes the similarity between the query $Q$ (i.e., the embedding of the flattened input image $z_t$) and the key $K$ (i.e., the embedding of each word $p^{(i)}$ in the prompt $p$). Therefore, the values in the attention map $M^{(i)}$ for a particular word $p^{(i)}$ indicate the similarity between the features of the corresponding region in the input image and the feature of word $p^{(i)}$. $V$ contains rich semantic features for each word $p^{(i)}$. According to the attention map, the cross-attention layer repeatedly adds the feature of $p^{(i)}$ to the area that is already similar enough to the word $p^{(i)}$. As a result, the generation of content related to a specific word $p^{(i)}$ ultimately occurs at the location with a strong signal in its attention map $M^{(i)}$.

\begin{figure*}[t]
  \centering
  \includegraphics[width=\linewidth]{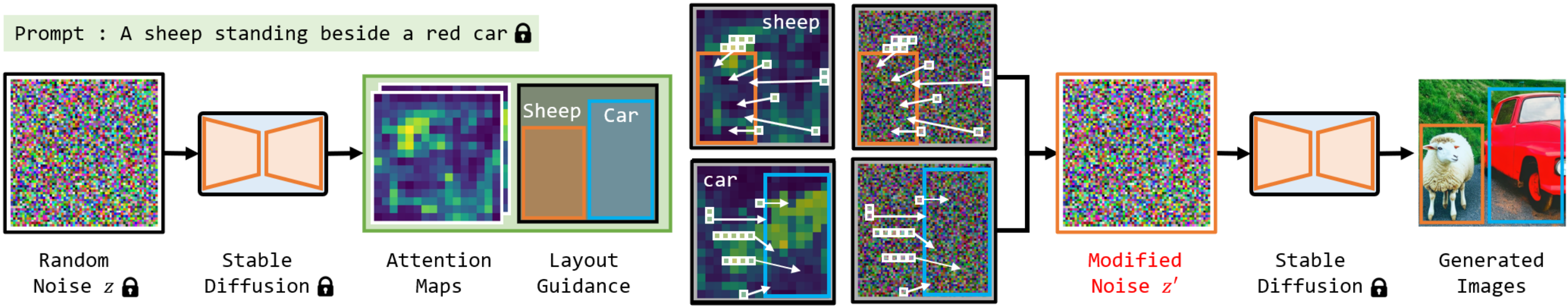}
  \vspace{-15pt}
  \caption{Concept of the pixel blocks swapping experiment. We use the attention map of the initial noise image to indicate the initial generation tendency. Subsequently, we move the pixel blocks that tend to generate specific content into specified regions, and the modified noise image is used to perform denoising as usual.}
  \vspace{-15pt}
  \label{fig:exp3}
\end{figure*}

\noindent\textbf{Image Synthesis from Random Initial Image} 
In the diffusion model, the initial images are typically sampled from a Gaussian distribution. The image generation process involves gradually moving the random starting point (initial image) towards the actual image distribution in the latent space. The diffusion model removes noise and utilizes a cross-attention layer to calculate the similarity between each pixel in the initial image and each object mentioned in the prompt. Then, the features of each object are assigned to the regions that already have similarities with that object. When a particular region in the initial noise image already contains features of a particular object mentioned in the prompt, the cross-attention layer adds more category features to improve the region towards a direction closer to that category, resulting in the generation of the object in that area. This mechanism constitutes the generation tendency of the initial image.

\noindent\textbf{Analysis of generation failure} According to our analysis, a critical factor contributing to the failure of image generation is the inconsistency between the initial noise image's generation tendency and the user-provided prompts. For instance, if a user intends to generate two objects close to each other, but the two regions that tend to generate these objects are far apart in the randomly generated initial image, the model will fail to generate the desired content based on such an initial image. Furthermore, some methods~\cite{balaji2022ediffi,mao2023trainingfree} aim to achieve layout-to-image generation by adding masks to the attention maps to increase the probability of generating specified objects in designated areas. However, if the user-specified regions on the initial image has a low tendency to generate the content, forcibly adding masks to such regions usually leads to failure.

\vspace{-10pt}
\subsection{Generated Image Re-painting}
\label{sec:exp2}

In this section, we investigate the impact of the partial changing of initial images on image generation in the diffusion model, which can be applied to re-painting tasks.

\noindent\textbf{To what extend can the initial noise image determine the content of the generated image?} As discussed in the previous chapter, the inconsistency between the generation tendency of the initialized image and the prompt is an essential factor leading to generation failure. Therefore, an intuitive idea to avoid image generation failure is to remove the part of the tendency that conflicts with the prompt. First, we generate an image using prompt $p$ and a randomly sampled initialized image $z$. We then identify regions in the generated image that did not match the description of $p$ or did not make sense. We re-randomize the regions in the initialized image $z$ corresponding to the failure regions while keeping the values of the other regions of $z$. We use the partially re-randomized initial noise image $z^{\prime}$ to perform the generation again, under the same prompt $p$, and observe the generated image. As shown in Fig.~\ref{fig:m2}, this experiment is straightforward and intuitive yet shows a significant effect. Using the partially re-randomized initial image $z^\prime$ achieves spontaneous regeneration of only the specified regions while keeping the other regions of the image almost unchanged. This result shows that the generation tendencies of each region in the initialized image are relatively independent, and changing a small part of the initialized image does not affect the generation tendencies of the remaining part. This experiment implies that by editing the initialized image, we can potentially influence the image generation results in a directional and controlled manner.

\noindent\textbf{Application on Generated Image Re-painting Task} A typical scenario in the diffusion model application scenario is that users have to perform dozens or even hundreds of generations to obtain a satisfactory image for a certain prompt. Each generation uses a freshly initial image sampled from a Gaussian distribution, resulting in each generation being independently randomized. However, users are sometimes satisfied with most of the content of a generated image while only dissatisfied with a small part of it. In this case, the user can only regenerate new images from the beginning. Our findings can be naturally applied to this scenario. When the user is satisfied with most of the content of a generated image (e.g., successfully generated objects, satisfactory layout, hue, and style, etc.), the user only needs to select the dissatisfied part and re-sample the corresponding pixel values in the initial image, and perform generation using the re-sampled image. The user can repeat this process until a satisfactory result is obtained  without worry about unexpected change on the other satisfied regions. This simple approach allows the user to obtain a satisfactory image faster. We present some re-painting samples in Fig.~\ref{fig:task_1}.

\vspace{-10pt}
\subsection{Layout-to-Image Synthesis}
\label{sec:exp3}
In this section, we apply our findings to the layout-to-image synthesis task and further explore the property of the generation tendency of the pixel blocks in the initial random image. Our previous experiments (Sec.~\ref{sec:exp1}) imply that the regions on the random initial images have the tendency to generate specific content. Further experiments (Sec.~\ref{sec:exp2}) have shown that changing the pixel values of the initial noisy image can correspondingly change the generated content at the exact location, implying that the generation tendencies of the regions in the initial noisy image are primarily determined by the values of the pixel points in them, rather than their locations. This finding can be applied to the layout-to-image synthesis task, where objects mentioned in the prompt are expected to be generated in user-specified regions.

\begin{figure}[t]
  \centering
  \includegraphics[width=\linewidth]{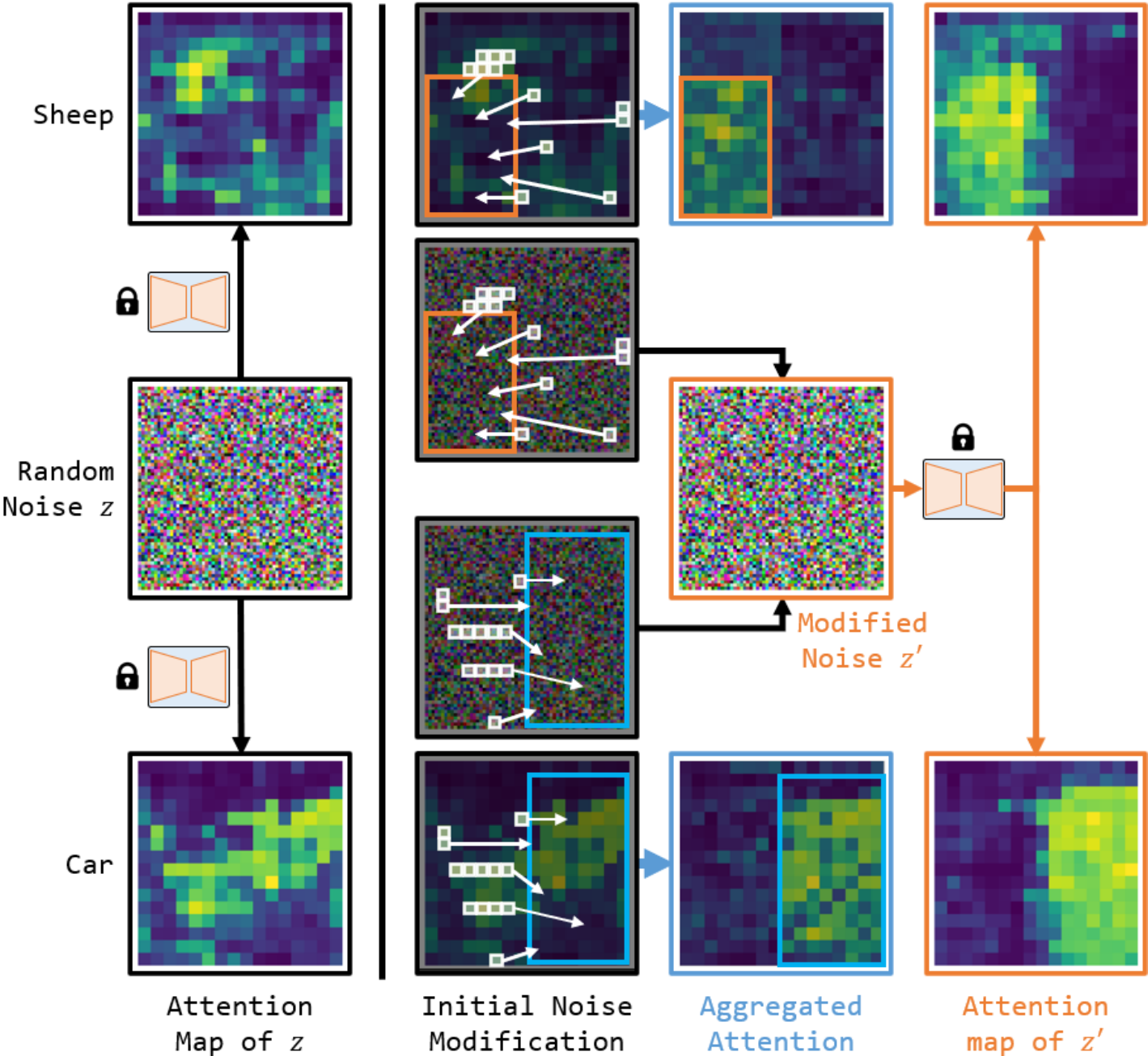}
  \vspace{-10pt}
  \caption{Attention maps after swapping and attention map calculated by modified initial noise $z^\prime$. Aggregated attention shows the swapped result of the original attention map. $z^\prime$ is obtained by swapping pixel blocks with high attention values into specified regions. The attention map of $z^\prime$ shows that the regions where high-attention pixel blocks are aggregated carry a high tendency to generate the corresponding content.}
  \label{fig:exp3_1}
  \vspace{-10pt}
\end{figure}

\noindent\textbf{Problem Setting}
In the layout-to-image generation task, not only the prompt $p$ is given, but the object-wise guidance $\{B, C\}$ is also provided, where $B$ is a list of specified regions, and $C$ is a list of categories. $\{b, c\} \in \{B, C\}$ means the object $c$ should be generated in region $b$. Each $c \in C$ should be mentioned in prompt $p$. Since we find that regions in the initial noise image have tendencies to generate certain categories, a straightforward and intuitive approach is to move all pixel blocks that tend to generate category $c$ to the specified region $b$. To achieve this, we need to know the generative tendencies of the pixel blocks in the initial noise image. 

\noindent\textbf{Generation Tendency of Pixel Blocks} As discussed in Sec.~\ref{sec:exp1}, the cross-attention map in U-Net calculates the generation tendency of each pixel block in current image. Therefore, we take the attention map from the first generation step of denoising to represent the generation tendency of pixel blocks in the initial noise image. Specifically, the size of the initial noise image used in Stable diffusion is $64 \times 64$. To preserve the structure and generation tendencies of the pixel blocks as much as possible, we take the attention map from the most down-sampled cross-attention layer, whose shape is $16 \times 16$, to represent the generation tendency of each pixel block in the initial image, i.e., each value in the attention map is corresponding to a $4 \times 4$ pixel block in the initial noise image. A pixel point with a significant value on the attention map of the word $c$ implies that a $4 \times 4$ pixel block in the initial image corresponding to this attention pixel has the tendency to generate the content $c$.

\noindent\textbf{Pixel Blocks Aggregation} For simplicity, we take the example of controlling the generated position $b$ of only one object $c$ in the following explanation, where $b$ is the corresponding area coordinated on the image re-scaled to $16 \times 16$. As discussed above, we can obtain the generation tendency of each pixel block in the initial noisy image based on the attention map. Next, we move the pixel blocks that tend to generate $c$, i.e., pixel blocks with high scores on the attention map of $c$, into the specified region $b$. Notably, our manipulation is performed on the initial image. We modify $z$ to $z^\prime$ according to the attention map and expect that using $z^\prime$ to perform the generation alone can achieve layout-to-image generation. Specifically, we calculate the area size $s$ of $b$, i.e., the number of attention pixels in the region $b$. We sort all the values on the attention map of $c$ from highest to lowest. Then we select the top $s$ of the sorted sequence of attention values. We use $A_{min}$ to denote the attention value of the $s$-th value we selected, which is the smallest one. We aim to move all selected attention values to the specified region $b$. Some of these values may already be located in the specified region. We keep the positions of these values unchanged and filter the coordinates in the specified region whose attention value is less than $A_{min}$ as follows,
\begin{equation}
    Out = \{p~|~attn(p) < A_{min} ,~p \in b\}
\end{equation}
and filter out the coordinates outside the specified region $b$ whose attention values are greater than $A_{max}$ as follows,
\begin{equation}
    In = \{p~|~attn(p) > A_{min} ,~p \notin b\}
\end{equation}
The number of elements in $In$ and $Out$ is the same. We randomly combine the elements of $In$ and $Out$ as swap pairs. Each element of $In$ and $Out$ is used and used only once.
\begin{equation}
    Swap = \{\{i,o\}~|~i \in In,~o \in Out \}
\end{equation}
Each position on the attention map corresponds to a $4 \times 4$ pixel block on the initial noise image, and for each swap pair $\{i,o\} \in Swap$, we swap the pixel blocks corresponding to $i$ and $o$ on the initial noise image. We swap the pixel blocks of $4 \times 4$ as invariant units, leaving their internal structure and values unchanged. The final generation is performed on the modified initial image $z^\prime$ by stable diffusion, as normal. The overall process is shown in Fig.~\ref{fig:exp3}. When regions of multiple objects $\{B, C\}$ are specified, to ensure that each pixel block is assigned to a unique object, we first classify these pixel blocks based on their values on the attention map of each category $c \in C$. For each $\{b, c\} \in \{B, C\}$, we sort and move the pixel blocks classified as $c$ to the corresponding region $b$.

\begin{table*}[t]
  \begin{center}
    \setlength{\tabcolsep}{4mm}{
  \begin{tabular}{l|cccc|cccc|c}
    \hline
     methods & IoU $\uparrow$  &  $\text{IoU}_s\uparrow$  & $\text{IoU}_m\uparrow$ & $\text{IoU}_l\uparrow$ & $R_{\text{suc}}\%\uparrow$ & $R^s_{\text{suc}}\%\uparrow$ &$R^m_{\text{suc}}\%\uparrow$ &$R^l_{\text{suc}}\%\uparrow$   & FID$\downarrow$  \\
    \hline
    \hline
    SD~\cite{rombach2022high} & $0.19$ & $0.05$ & $0.21$ & $0.38$ & $12.14$ &$ 0.50$ & $8.11$ & $39.41$ & $20.8$ \\
    Paint~\cite{balaji2022ediffi}  &  $0.24$ & $0.08$ & $0.27$ & $0.46$ & $17.56$ & $1.16$ & $14.05$ & $52.34$ & $22.7$ \\
    Soft~\cite{mao2023trainingfree} & $0.27$ &  $0.10$  & $0.31$  &  $0.48$ &   $22.31$ &  $2.95$  &  $22.38$ &  $56.24$ &  $21.3$\\
    \hline
    Ours & $0.26$ & $0.09$ &  $0.32$ & $0.46$ & $21.58$ & $2.26$ &  $22.75$ & $54.21$ & $21.7$\\
    Paint~\cite{balaji2022ediffi} + Ours & $0.29$ & $0.11$ & $0.35$ & $0.51$ & $23.89$ & $2.12$ & $24.74$ & $61.48$ & $23.7$\\
    Soft~\cite{mao2023trainingfree} + Ours & \bm{$0.31$} & \bm{$0.12$} & \bm{$0.37$} & \bm{$0.52$} & \bm{$26.41$} & \bm{$3.24$} & \bm{$29.55$} & \bm{$62.59$} & \bm{$21.2$}\\
    \hline
  \end{tabular}}
\end{center}

\caption{Control effectiveness of layout guidance. Our method achieves a comparable performance with state-of-the-art methods when used alone, and combining our method with them achieves the best performance on all subsets.}
\vspace{-20pt}
\label{tab:results}
\end{table*}

\noindent\textbf{Generation tendency of Pixel Blocks After Moving} As shown in Fig.~\ref{fig:exp3_1}, we move the pixel blocks in the initial image $z$ that tend to generate content $c$ into the specified region $b$ and obtain modified initial image $z^\prime$. We subsequently obtain the attention map of $z^\prime$ through the U-Net of the same stable diffusion model. We observe that the specified region $b$ in the modified image $z^\prime$ shows a clear tendency to generate category $c$, as shown by the high-intensity signal in the corresponding region $b$ on the attention map of category $c$. This phenomenon means that the generative tendencies of these pixel blocks follow the movement of the values, further indicating that the generative tendency of a pixel block is mainly determined by the pixel values rather than its spatial location.

\noindent\textbf{Comparing with Mask-based Methods} Some methods~\cite{balaji2022ediffi, mao2023trainingfree} try to increase the probability of the specified category appearing on the specified region by artificially adding a mask to the attention map of the specified category. This approach and our approach both result in higher values in the attention map of the specified region eventually. However, mask-based methods are fundamentally limited by the fact that they entirely ignore the generation tendency of the initial image itself. If the initial pixels in the specified region are inherently hard to generate the specified class, the mask-based approach of forcefully adding features of the specified class to the specified region is likely to cause generation failure or affect the quality of the image. On the other hand, if other regions outside the specified region have a tendency to generate the specified category inherently, even adding a mask to the specified region can not prevent the generation of objects outside the region. Unlike these methods, shifting the initialized image pixel blocks fundamentally modifies the initialized image's generative tendency. The model will spontaneously generate the specified objects in the specified regions in subsequent generations, even if no additional interference is implied.

\vspace{-10pt}
\section{Evaluation}

To confirm the effectiveness of our proposal in the layout-to-image synthesis task and to demonstrate the advantages of our approach, we quantitatively evaluate our method and compare it with state-of-the-art methods. We also conducted experiments to boost the performance of mask-based methods by combining them with our method. The experimental setup follows Soft-Mask~\cite{mao2023trainingfree}, where an object detector is utilized to evaluate the efficiency of object-wise controlling.

\vspace{-10pt}
\subsection{Dataset}
We conduct a quantitative evaluation on the MS COCO~\cite{lin2014microsoft} dataset, which contains image caption and object annotation. We selected samples whose bounding box categories are all mentioned in their image captions. Since most of the image captions involve only one or two significant objects in the image, we randomly selected 6000 images, where 3000 contain one object, and the other 3000 contain two, as in the previous work. All images and their corresponding annotations are resized to $512^2$. During denoising, the captions of images are used as prompts, and the object annotations are used as layout guidance.

\begin{figure*}[t]
  \centering
  \includegraphics[width=0.98\linewidth]{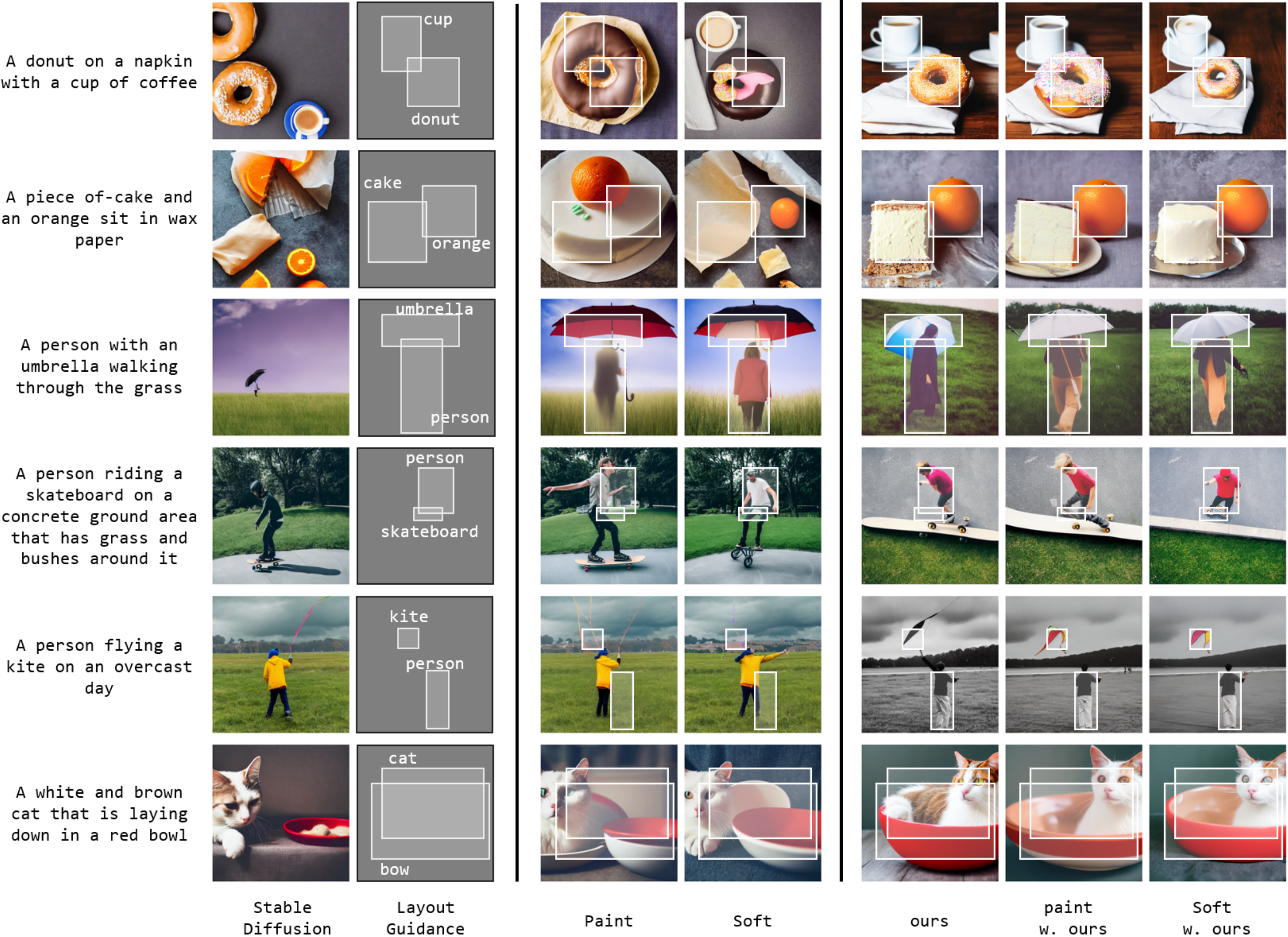}
  \vspace{-10pt}
  \caption{Quality samples in our experiments on MS COCO. Images on each row are generated from the same initial noise image. Images without applying our method have significant commonalities (structure, background, style) in most cases, even if different masks are applied. When our method is applied, some positions of pixel blocks are swapped, and the initial images are severely changed. Thus, the generated images are accordingly changed.}
  \vspace{-10pt}
  \label{fig:vis}
\end{figure*}

\vspace{-10pt}
\subsection{Metric}
\noindent\textbf{Efficacy of Layout Controlling} We generate images using image captions as prompts and object-wise annotation as layout guidance. Then, we use an object detector, YOLOR~\cite{wang2021you}, to perform object detection on generated images and obtain the categories and detected bounding boxes of generated objects. Then the consistency of the detection results with the layout guidance can represent the effectiveness of layout controlling. For each object, the IoU between its layout guidance (i.e., bounding box annotation in the dataset) and detected bounding box is calculated. If the IoU is larger than $0.5$, it is regarded as a successful control, and the success rate is represented by $R_\text{suc}$. To further evaluate the effectiveness of objects of different sizes, all objects are categorized into several subsets according to their size and distance from the image center. Specifically, the subset $s$ contains objects with an area less than $150^2$, the subset $l$ contains all objects with an area greater than $300^2$, while the other objects are channeled into subset $m$. 

%% According to the distances between the guidance bounding boxes and the center of the image, objects are equally divided into three subsets, $near$, $mid$, and $far$.

\noindent\textbf{Image Quality} FID~\cite{heusel2017gans} measures the similarity between two groups of images and is commonly used to indicate the quality of generated images. We calculate the FID by generated images and ground truth images (i.e., images in the dataset). A smaller FID indicates a better image quality.

\vspace{-10pt}
\subsection{Implementation}
We utilize the stable diffusion (SD)~\cite{rombach2022high} v1.4 with the default configuration as the baseline for our study. In this setting, the model undergoes 50 denoising sampling steps using PLMS. The baseline stable diffusion do not receive any layout guidance, but may by chance generate the specified objects in the specified regions. During each image generation process in other experiments, the model receives both prompt and layout guidance and produced corresponding images. Our method uses the model's U-Net to compute the cross-attention map based on the randomly sampled initial image and prompt ${z,p}$. Subsequently, we move pixel blocks in $z$ with high attention scores to the designated area by swapping to obtain the modified initial noise image $z^\prime$. We then perform normal denoising using $\{z^\prime,p\}$. We compare our approach to Paint~\cite{balaji2022ediffi} and Soft~\cite{mao2023trainingfree}. They both add masks to user-specified regions to enhance attention values in the corresponding regions. Soft~\cite {mao2023trainingfree} additionally suppresses the out-of-region attention. We use the same $\{z,p\}$ for denoising and add the layout mask to the cross-attention map of each U-Net layer at every denoising step. Additionally, we report the performance when our method is combined with mask-based methods. In this scenario, we generate images using $\{z^\prime,p\}$ and employ masks during denoising. 

\vspace{-10pt}
\subsection{Results and Analysis}
Evaluation results are shown in Tab.~\ref{tab:results}. Our method significantly outperform the baseline~\cite{rombach2022high} and Paint~\cite{balaji2022ediffi} on all subsets, and achieves comparable performance with Soft~\cite{mao2023trainingfree}, without applying any interference to the denoising process of the model. Additionally, our method and mask-based methods function at different stages of generation. Considering the denoising process as moving a random starting point in the latent space to a target point, our method can be viewed as choosing a better starting point for the denoising process. At the same time, mask-based methods can be considered as providing an additional guide toward a fixed direction during the movement. Thus the two methods are not in conflict with each other but can be used together naturally. Intuitively, combining with our method significantly improves the performance of both mask-based methods, while the combination of ours and Soft~\cite{mao2023trainingfree} achieves the best performance on all subsets. Some quality samples are shown in Fig.~\ref{fig:vis}. The effectiveness of our method demonstrates that the generation tendency of the initial random image mainly depends on pixel values rather than spatial location, and the generation tendency follows the moving of pixel values on the initial image. As a limitation of our method, when the guidance bounding box is small, our method moves only a few pixels in the initial image and sometimes cannot significantly affect the generation. Therefore, the effectiveness of our method is relatively weak for small objects compared to larger ones.

\vspace{-10pt}
\section{Conclusion and Future Works}
In this paper, we present a novel direction for fine-grained control in image generation by diffusion model that is to modify the random sampled initial noise image. Although the initial noise image has traditionally been considered a random noise signal lacking any meaningful information, our research has uncovered that such an image possesses an inherent inclination to generate particular content at specific locations when used as a starting point for denoising. We investigate the underlying factors contributing to this tendency and endeavor to adjust the initial image's generation tendency to impact the final output. Our results demonstrate that altering a particular section of the initial noise image does affect the corresponding content in the generated image, which we leverage in the repainting task. Moreover, we uncover that this generation tendency primarily hinges on pixel values rather than spatial positioning, shifting with the motion of pixel values in the initial image. We apply this discovery to the layout-to-image generation task and achieve state-of-the-art performances. Based on the findings in our paper, several exciting potential research directions have emerged. Specifically, Our approach highlights the significance of the initial image in the image generation process. Therefore, research on optimizing the initial image or accelerating denoising based on the optimized initial image holds excellent promise. 

\noindent\textbf{Acknowledgement} 

\noindent This research was partially supported by JST Mirai-Program JPMJMI21H1.

\bibliographystyle{ACM-Reference-Format}
\balance
\bibliography{ref}


\begin{thebibliography}{32}

%%% the Web address.

\ifx \showCODEN    \undefined \def \showCODEN     #1{\unskip}     \fi
\ifx \showDOI      \undefined \def \showDOI       #1{#1}\fi
\ifx \showISBNx    \undefined \def \showISBNx     #1{\unskip}     \fi
\ifx \showISBNxiii \undefined \def \showISBNxiii  #1{\unskip}     \fi
\ifx \showISSN     \undefined \def \showISSN      #1{\unskip}     \fi
\ifx \showLCCN     \undefined \def \showLCCN      #1{\unskip}     \fi
\ifx \shownote     \undefined \def \shownote      #1{#1}          \fi
\ifx \showarticletitle \undefined \def \showarticletitle #1{#1}   \fi
\ifx \showURL      \undefined \def \showURL       {\relax}        \fi
% invisible to TeX
\providecommand\bibfield[2]{#2}
\providecommand\bibinfo[2]{#2}
\providecommand\natexlab[1]{#1}
\providecommand\showeprint[2][]{arXiv:#2}

\bibitem[Avrahami et~al\mbox{.}(2022)]%
        {avrahami2022spatext}
\bibfield{author}{\bibinfo{person}{Omri Avrahami}, \bibinfo{person}{Thomas Hayes}, \bibinfo{person}{Oran Gafni}, \bibinfo{person}{Sonal Gupta}, \bibinfo{person}{Yaniv Taigman}, \bibinfo{person}{Devi Parikh}, \bibinfo{person}{Dani Lischinski}, \bibinfo{person}{Ohad Fried}, {and} \bibinfo{person}{Xi Yin}.} \bibinfo{year}{2022}\natexlab{}.
\newblock \showarticletitle{SpaText: Spatio-Textual Representation for Controllable Image Generation}.
\newblock \bibinfo{journal}{\emph{arXiv preprint arXiv:2211.14305}} (\bibinfo{year}{2022}).
\newblock


\bibitem[Balaji et~al\mbox{.}(2022)]%
        {balaji2022ediffi}
\bibfield{author}{\bibinfo{person}{Yogesh Balaji}, \bibinfo{person}{Seungjun Nah}, \bibinfo{person}{Xun Huang}, \bibinfo{person}{Arash Vahdat}, \bibinfo{person}{Jiaming Song}, \bibinfo{person}{Karsten Kreis}, \bibinfo{person}{Miika Aittala}, \bibinfo{person}{Timo Aila}, \bibinfo{person}{Samuli Laine}, \bibinfo{person}{Bryan Catanzaro}, {et~al\mbox{.}}} \bibinfo{year}{2022}\natexlab{}.
\newblock \showarticletitle{ediffi: Text-to-image diffusion models with an ensemble of expert denoisers}.
\newblock \bibinfo{journal}{\emph{arXiv preprint arXiv:2211.01324}} (\bibinfo{year}{2022}).
\newblock


\bibitem[Dhariwal and Nichol(2021)]%
        {dhariwal2021diffusion}
\bibfield{author}{\bibinfo{person}{Prafulla Dhariwal} {and} \bibinfo{person}{Alexander Nichol}.} \bibinfo{year}{2021}\natexlab{}.
\newblock \showarticletitle{Diffusion models beat gans on image synthesis}.
\newblock \bibinfo{journal}{\emph{NeurIPS}}  \bibinfo{volume}{34} (\bibinfo{year}{2021}), \bibinfo{pages}{8780--8794}.
\newblock


\bibitem[Ding et~al\mbox{.}(2021)]%
        {ding2021cogview}
\bibfield{author}{\bibinfo{person}{Ming Ding}, \bibinfo{person}{Zhuoyi Yang}, \bibinfo{person}{Wenyi Hong}, \bibinfo{person}{Wendi Zheng}, \bibinfo{person}{Chang Zhou}, \bibinfo{person}{Da Yin}, \bibinfo{person}{Junyang Lin}, \bibinfo{person}{Xu Zou}, \bibinfo{person}{Zhou Shao}, \bibinfo{person}{Hongxia Yang}, {et~al\mbox{.}}} \bibinfo{year}{2021}\natexlab{}.
\newblock \showarticletitle{Cogview: Mastering text-to-image generation via transformers}.
\newblock \bibinfo{journal}{\emph{NeurIPS}}  \bibinfo{volume}{34} (\bibinfo{year}{2021}), \bibinfo{pages}{19822--19835}.
\newblock


\bibitem[Feng et~al\mbox{.}(2022)]%
        {feng2022training}
\bibfield{author}{\bibinfo{person}{Weixi Feng}, \bibinfo{person}{Xuehai He}, \bibinfo{person}{Tsu-Jui Fu}, \bibinfo{person}{Varun Jampani}, \bibinfo{person}{Arjun Akula}, \bibinfo{person}{Pradyumna Narayana}, \bibinfo{person}{Sugato Basu}, \bibinfo{person}{Xin~Eric Wang}, {and} \bibinfo{person}{William~Yang Wang}.} \bibinfo{year}{2022}\natexlab{}.
\newblock \showarticletitle{Training-Free Structured Diffusion Guidance for Compositional Text-to-Image Synthesis}.
\newblock \bibinfo{journal}{\emph{arXiv preprint arXiv:2212.05032}} (\bibinfo{year}{2022}).
\newblock


\bibitem[Gafni et~al\mbox{.}(2022)]%
        {gafni2022make}
\bibfield{author}{\bibinfo{person}{Oran Gafni}, \bibinfo{person}{Adam Polyak}, \bibinfo{person}{Oron Ashual}, \bibinfo{person}{Shelly Sheynin}, \bibinfo{person}{Devi Parikh}, {and} \bibinfo{person}{Yaniv Taigman}.} \bibinfo{year}{2022}\natexlab{}.
\newblock \showarticletitle{Make-a-scene: Scene-based text-to-image generation with human priors}. In \bibinfo{booktitle}{\emph{ECCV}}. Springer, \bibinfo{pages}{89--106}.
\newblock


\bibitem[Hertz et~al\mbox{.}(2022)]%
        {hertz2022prompt}
\bibfield{author}{\bibinfo{person}{Amir Hertz}, \bibinfo{person}{Ron Mokady}, \bibinfo{person}{Jay Tenenbaum}, \bibinfo{person}{Kfir Aberman}, \bibinfo{person}{Yael Pritch}, {and} \bibinfo{person}{Daniel Cohen-Or}.} \bibinfo{year}{2022}\natexlab{}.
\newblock \showarticletitle{Prompt-to-prompt image editing with cross attention control}.
\newblock \bibinfo{journal}{\emph{arXiv preprint arXiv:2208.01626}} (\bibinfo{year}{2022}).
\newblock


\bibitem[Heusel et~al\mbox{.}(2017)]%
        {heusel2017gans}
\bibfield{author}{\bibinfo{person}{Martin Heusel}, \bibinfo{person}{Hubert Ramsauer}, \bibinfo{person}{Thomas Unterthiner}, \bibinfo{person}{Bernhard Nessler}, {and} \bibinfo{person}{Sepp Hochreiter}.} \bibinfo{year}{2017}\natexlab{}.
\newblock \showarticletitle{Gans trained by a two time-scale update rule converge to a local nash equilibrium}.
\newblock \bibinfo{journal}{\emph{NeurIPS}}  \bibinfo{volume}{30} (\bibinfo{year}{2017}).
\newblock


\bibitem[Ho et~al\mbox{.}(2020)]%
        {ho2020denoising}
\bibfield{author}{\bibinfo{person}{Jonathan Ho}, \bibinfo{person}{Ajay Jain}, {and} \bibinfo{person}{Pieter Abbeel}.} \bibinfo{year}{2020}\natexlab{}.
\newblock \showarticletitle{Denoising diffusion probabilistic models}.
\newblock \bibinfo{journal}{\emph{NeurIPS}}  \bibinfo{volume}{33} (\bibinfo{year}{2020}), \bibinfo{pages}{6840--6851}.
\newblock


\bibitem[Ho and Salimans({[n.\,d.]})]%
        {ho2021classifier}
\bibfield{author}{\bibinfo{person}{Jonathan Ho} {and} \bibinfo{person}{Tim Salimans}.} \bibinfo{year}{[n.\,d.]}\natexlab{}.
\newblock \showarticletitle{Classifier-Free Diffusion Guidance}. In \bibinfo{booktitle}{\emph{NeurIPS 2021 Workshop on Deep Generative Models and Downstream Applications}}.
\newblock


\bibitem[Huang et~al\mbox{.}(2023)]%
        {huang2023composer}
\bibfield{author}{\bibinfo{person}{Lianghua Huang}, \bibinfo{person}{Di Chen}, \bibinfo{person}{Yu Liu}, \bibinfo{person}{Yujun Shen}, \bibinfo{person}{Deli Zhao}, {and} \bibinfo{person}{Jingren Zhou}.} \bibinfo{year}{2023}\natexlab{}.
\newblock \showarticletitle{Composer: Creative and controllable image synthesis with composable conditions}.
\newblock \bibinfo{journal}{\emph{arXiv preprint arXiv:2302.09778}} (\bibinfo{year}{2023}).
\newblock


\bibitem[Kawar et~al\mbox{.}(2022)]%
        {kawar2022imagic}
\bibfield{author}{\bibinfo{person}{Bahjat Kawar}, \bibinfo{person}{Shiran Zada}, \bibinfo{person}{Oran Lang}, \bibinfo{person}{Omer Tov}, \bibinfo{person}{Huiwen Chang}, \bibinfo{person}{Tali Dekel}, \bibinfo{person}{Inbar Mosseri}, {and} \bibinfo{person}{Michal Irani}.} \bibinfo{year}{2022}\natexlab{}.
\newblock \showarticletitle{Imagic: Text-based real image editing with diffusion models}.
\newblock \bibinfo{journal}{\emph{arXiv preprint arXiv:2210.09276}} (\bibinfo{year}{2022}).
\newblock


\bibitem[Kim et~al\mbox{.}(2022)]%
        {kim2022diffusionclip}
\bibfield{author}{\bibinfo{person}{Gwanghyun Kim}, \bibinfo{person}{Taesung Kwon}, {and} \bibinfo{person}{Jong~Chul Ye}.} \bibinfo{year}{2022}\natexlab{}.
\newblock \showarticletitle{Diffusionclip: Text-guided diffusion models for robust image manipulation}. In \bibinfo{booktitle}{\emph{CVPR}}. \bibinfo{pages}{2426--2435}.
\newblock


\bibitem[Lin et~al\mbox{.}(2014)]%
        {lin2014microsoft}
\bibfield{author}{\bibinfo{person}{Tsung-Yi Lin}, \bibinfo{person}{Michael Maire}, \bibinfo{person}{Serge Belongie}, \bibinfo{person}{James Hays}, \bibinfo{person}{Pietro Perona}, \bibinfo{person}{Deva Ramanan}, \bibinfo{person}{Piotr Doll{\'a}r}, {and} \bibinfo{person}{C~Lawrence Zitnick}.} \bibinfo{year}{2014}\natexlab{}.
\newblock \showarticletitle{Microsoft coco: Common objects in context}. In \bibinfo{booktitle}{\emph{ECCV}}. Springer, \bibinfo{pages}{740--755}.
\newblock


\bibitem[Liu et~al\mbox{.}(2022b)]%
        {liu2022pseudo}
\bibfield{author}{\bibinfo{person}{Luping Liu}, \bibinfo{person}{Yi Ren}, \bibinfo{person}{Zhijie Lin}, {and} \bibinfo{person}{Zhou Zhao}.} \bibinfo{year}{2022}\natexlab{b}.
\newblock \showarticletitle{Pseudo Numerical Methods for Diffusion Models on Manifolds}. In \bibinfo{booktitle}{\emph{ICLR}}.
\newblock


\bibitem[Liu et~al\mbox{.}(2022a)]%
        {liu2022compositional}
\bibfield{author}{\bibinfo{person}{Nan Liu}, \bibinfo{person}{Shuang Li}, \bibinfo{person}{Yilun Du}, \bibinfo{person}{Antonio Torralba}, {and} \bibinfo{person}{Joshua~B Tenenbaum}.} \bibinfo{year}{2022}\natexlab{a}.
\newblock \showarticletitle{Compositional visual generation with composable diffusion models}. In \bibinfo{booktitle}{\emph{ECCV}}. Springer, \bibinfo{pages}{423--439}.
\newblock


\bibitem[Mao and Wang(2023)]%
        {mao2023trainingfree}
\bibfield{author}{\bibinfo{person}{Jiafeng Mao} {and} \bibinfo{person}{Xueting Wang}.} \bibinfo{year}{2023}\natexlab{}.
\newblock \showarticletitle{Training-Free Location-Aware Text-to-Image Synthesis}.
\newblock \bibinfo{journal}{\emph{arXiv preprint arXiv:2304.13427}} (\bibinfo{year}{2023}).
\newblock


\bibitem[Mou et~al\mbox{.}(2023)]%
        {mou2023t2i}
\bibfield{author}{\bibinfo{person}{Chong Mou}, \bibinfo{person}{Xintao Wang}, \bibinfo{person}{Liangbin Xie}, \bibinfo{person}{Jian Zhang}, \bibinfo{person}{Zhongang Qi}, \bibinfo{person}{Ying Shan}, {and} \bibinfo{person}{Xiaohu Qie}.} \bibinfo{year}{2023}\natexlab{}.
\newblock \showarticletitle{T2I-Adapter: Learning Adapters to Dig out More Controllable Ability for Text-to-Image Diffusion Models}.
\newblock \bibinfo{journal}{\emph{arXiv e-prints}} (\bibinfo{year}{2023}), \bibinfo{pages}{arXiv--2302}.
\newblock


\bibitem[Nichol and Dhariwal(2021)]%
        {nichol2021improved}
\bibfield{author}{\bibinfo{person}{Alexander~Quinn Nichol} {and} \bibinfo{person}{Prafulla Dhariwal}.} \bibinfo{year}{2021}\natexlab{}.
\newblock \showarticletitle{Improved denoising diffusion probabilistic models}. In \bibinfo{booktitle}{\emph{ICML}}. PMLR, \bibinfo{pages}{8162--8171}.
\newblock


\bibitem[Nichol et~al\mbox{.}(2022)]%
        {nichol2022glide}
\bibfield{author}{\bibinfo{person}{Alexander~Quinn Nichol}, \bibinfo{person}{Prafulla Dhariwal}, \bibinfo{person}{Aditya Ramesh}, \bibinfo{person}{Pranav Shyam}, \bibinfo{person}{Pamela Mishkin}, \bibinfo{person}{Bob Mcgrew}, \bibinfo{person}{Ilya Sutskever}, {and} \bibinfo{person}{Mark Chen}.} \bibinfo{year}{2022}\natexlab{}.
\newblock \showarticletitle{GLIDE: Towards Photorealistic Image Generation and Editing with Text-Guided Diffusion Models}. In \bibinfo{booktitle}{\emph{ICML}}. PMLR, \bibinfo{pages}{16784--16804}.
\newblock


\bibitem[Park et~al\mbox{.}(2021)]%
        {park2021benchmark}
\bibfield{author}{\bibinfo{person}{Dong~Huk Park}, \bibinfo{person}{Samaneh Azadi}, \bibinfo{person}{Xihui Liu}, \bibinfo{person}{Trevor Darrell}, {and} \bibinfo{person}{Anna Rohrbach}.} \bibinfo{year}{2021}\natexlab{}.
\newblock \showarticletitle{Benchmark for compositional text-to-image synthesis}. In \bibinfo{booktitle}{\emph{NeurIPS Datasets and Benchmarks Track (Round 1)}}.
\newblock


\bibitem[Radford et~al\mbox{.}(2021)]%
        {radford2021learning}
\bibfield{author}{\bibinfo{person}{Alec Radford}, \bibinfo{person}{Jong~Wook Kim}, \bibinfo{person}{Chris Hallacy}, \bibinfo{person}{Aditya Ramesh}, \bibinfo{person}{Gabriel Goh}, \bibinfo{person}{Sandhini Agarwal}, \bibinfo{person}{Girish Sastry}, \bibinfo{person}{Amanda Askell}, \bibinfo{person}{Pamela Mishkin}, \bibinfo{person}{Jack Clark}, {et~al\mbox{.}}} \bibinfo{year}{2021}\natexlab{}.
\newblock \showarticletitle{Learning transferable visual models from natural language supervision}. In \bibinfo{booktitle}{\emph{ICML}}. PMLR, \bibinfo{pages}{8748--8763}.
\newblock


\bibitem[Ramesh et~al\mbox{.}(2022)]%
        {ramesh2022hierarchical}
\bibfield{author}{\bibinfo{person}{Aditya Ramesh}, \bibinfo{person}{Prafulla Dhariwal}, \bibinfo{person}{Alex Nichol}, \bibinfo{person}{Casey Chu}, {and} \bibinfo{person}{Mark Chen}.} \bibinfo{year}{2022}\natexlab{}.
\newblock \showarticletitle{Hierarchical text-conditional image generation with clip latents}.
\newblock \bibinfo{journal}{\emph{arXiv preprint arXiv:2204.06125}} (\bibinfo{year}{2022}).
\newblock


\bibitem[Ramesh et~al\mbox{.}(2021)]%
        {ramesh2021zero}
\bibfield{author}{\bibinfo{person}{Aditya Ramesh}, \bibinfo{person}{Mikhail Pavlov}, \bibinfo{person}{Gabriel Goh}, \bibinfo{person}{Scott Gray}, \bibinfo{person}{Chelsea Voss}, \bibinfo{person}{Alec Radford}, \bibinfo{person}{Mark Chen}, {and} \bibinfo{person}{Ilya Sutskever}.} \bibinfo{year}{2021}\natexlab{}.
\newblock \showarticletitle{Zero-shot text-to-image generation}. In \bibinfo{booktitle}{\emph{ICML}}. PMLR, \bibinfo{pages}{8821--8831}.
\newblock


\bibitem[Rombach et~al\mbox{.}(2022)]%
        {rombach2022high}
\bibfield{author}{\bibinfo{person}{Robin Rombach}, \bibinfo{person}{Andreas Blattmann}, \bibinfo{person}{Dominik Lorenz}, \bibinfo{person}{Patrick Esser}, {and} \bibinfo{person}{Bj{\"o}rn Ommer}.} \bibinfo{year}{2022}\natexlab{}.
\newblock \showarticletitle{High-resolution image synthesis with latent diffusion models}. In \bibinfo{booktitle}{\emph{CVPR}}. \bibinfo{pages}{10684--10695}.
\newblock


\bibitem[Ronneberger et~al\mbox{.}(2015)]%
        {ronneberger2015u}
\bibfield{author}{\bibinfo{person}{Olaf Ronneberger}, \bibinfo{person}{Philipp Fischer}, {and} \bibinfo{person}{Thomas Brox}.} \bibinfo{year}{2015}\natexlab{}.
\newblock \showarticletitle{U-net: Convolutional networks for biomedical image segmentation}. In \bibinfo{booktitle}{\emph{MICCAI}}. Springer, \bibinfo{pages}{234--241}.
\newblock


\bibitem[Song et~al\mbox{.}(2021)]%
        {song2020denoising}
\bibfield{author}{\bibinfo{person}{Jiaming Song}, \bibinfo{person}{Chenlin Meng}, {and} \bibinfo{person}{Stefano Ermon}.} \bibinfo{year}{2021}\natexlab{}.
\newblock \showarticletitle{Denoising Diffusion Implicit Models}. In \bibinfo{booktitle}{\emph{ICLR}}.
\newblock


\bibitem[Vaswani et~al\mbox{.}(2017)]%
        {vaswani2017attention}
\bibfield{author}{\bibinfo{person}{Ashish Vaswani}, \bibinfo{person}{Noam Shazeer}, \bibinfo{person}{Niki Parmar}, \bibinfo{person}{Jakob Uszkoreit}, \bibinfo{person}{Llion Jones}, \bibinfo{person}{Aidan~N Gomez}, \bibinfo{person}{{\L}ukasz Kaiser}, {and} \bibinfo{person}{Illia Polosukhin}.} \bibinfo{year}{2017}\natexlab{}.
\newblock \showarticletitle{Attention is all you need}.
\newblock \bibinfo{journal}{\emph{NeurIPS}}  \bibinfo{volume}{30} (\bibinfo{year}{2017}).
\newblock


\bibitem[Voynov et~al\mbox{.}(2022)]%
        {voynov2022sketch}
\bibfield{author}{\bibinfo{person}{Andrey Voynov}, \bibinfo{person}{Kfir Aberman}, {and} \bibinfo{person}{Daniel Cohen-Or}.} \bibinfo{year}{2022}\natexlab{}.
\newblock \showarticletitle{Sketch-Guided Text-to-Image Diffusion Models}.
\newblock \bibinfo{journal}{\emph{arXiv preprint arXiv:2211.13752}} (\bibinfo{year}{2022}).
\newblock


\bibitem[Wang et~al\mbox{.}(2021)]%
        {wang2021you}
\bibfield{author}{\bibinfo{person}{Chien-Yao Wang}, \bibinfo{person}{I-Hau Yeh}, {and} \bibinfo{person}{Hong-Yuan~Mark Liao}.} \bibinfo{year}{2021}\natexlab{}.
\newblock \showarticletitle{You only learn one representation: Unified network for multiple tasks}.
\newblock \bibinfo{journal}{\emph{arXiv preprint arXiv:2105.04206}} (\bibinfo{year}{2021}).
\newblock


\bibitem[Wang et~al\mbox{.}(2022)]%
        {wang2022pretraining}
\bibfield{author}{\bibinfo{person}{Tengfei Wang}, \bibinfo{person}{Ting Zhang}, \bibinfo{person}{Bo Zhang}, \bibinfo{person}{Hao Ouyang}, \bibinfo{person}{Dong Chen}, \bibinfo{person}{Qifeng Chen}, {and} \bibinfo{person}{Fang Wen}.} \bibinfo{year}{2022}\natexlab{}.
\newblock \showarticletitle{Pretraining is all you need for image-to-image translation}.
\newblock \bibinfo{journal}{\emph{arXiv preprint arXiv:2205.12952}} (\bibinfo{year}{2022}).
\newblock


\bibitem[Zhang and Agrawala(2023)]%
        {zhang2023adding}
\bibfield{author}{\bibinfo{person}{Lvmin Zhang} {and} \bibinfo{person}{Maneesh Agrawala}.} \bibinfo{year}{2023}\natexlab{}.
\newblock \showarticletitle{Adding conditional control to text-to-image diffusion models}.
\newblock \bibinfo{journal}{\emph{arXiv preprint arXiv:2302.05543}} (\bibinfo{year}{2023}).
\newblock


\end{thebibliography}

\end{document}